\documentclass[letterpaper, 10pt, conference]{IEEEtran}
\IEEEoverridecommandlockouts

\usepackage[font=small]{caption}
\usepackage[T1]{fontenc}
\usepackage{comment}
\usepackage{amssymb}
\usepackage{graphicx}
\usepackage{mathtools}
\usepackage{amsmath}
\usepackage{gensymb}
\usepackage{float}
\usepackage{multirow}
\usepackage{makecell}
\usepackage{mathtools}
\usepackage{wrapfig}
\usepackage{siunitx}
\usepackage{overpic}
\usepackage{subcaption}
\usepackage{amsmath}
\usepackage{titlesec}
\usepackage{lipsum}

\usepackage{booktabs}
\usepackage{multirow}
\usepackage{siunitx}

\usepackage{tabularx}
    \newcolumntype{L}{>{\raggedright\arraybackslash}X}
\usepackage[utf8]{inputenc}
\usepackage[english]{babel}
\usepackage{amsthm}
\usepackage{algpseudocode}
\usepackage[dvipsnames, table]{xcolor}
\usepackage{bm}
\usepackage[mathscr]{eucal}
\usepackage{bm}
\usepackage{nicefrac}
\usepackage[ruled,vlined,linesnumbered]{algorithm2e}
\usepackage{makecell}
\usepackage{graphicx}
\usepackage{caption}
\usepackage{multirow}
\usepackage{makecell}
\usepackage{booktabs}
\usepackage{dsfont}
\usepackage[textsize=tiny]{todonotes}

\usepackage{xspace}
\usepackage{subcaption}

\newcommand{\model}{\textsc{FastFunnels}\xspace}

\usepackage{parskip}

\definecolor{ff}{HTML}{1F77B4}

 \title{\LARGE \bf \textsc{FastFunnels}: Agile Multi Robot Navigation in Narrow Spaces via Adaptive Dynamic Funnels}
\author{Vignesh Rajagopal \quad Rohan Chandra \\
}

\makeatletter
\let\NAT@parse\undefined
\makeatother
\usepackage[dvipsnames]{xcolor}
\usepackage[colorlink
s=true,citecolor=orange,linkcolor=red, urlcolor=magenta]{hyperref}
\usepackage{bm}
\usepackage{wrapfig}
\usepackage{booktabs}
\usepackage{graphicx}
\usepackage{enumitem}
\usepackage{cite}
\usepackage[table]{xcolor}   % the [table] option enables \rowcolor
\usepackage{multirow}

\begin{document}
\raggedbottom

\maketitle
\thispagestyle{plain}
\pagestyle{plain}

% comment 1 - \rc{youre presenting an algorithm, not a framework.}
% comment 2 - \rc{``performance is vague - be specific about what you are improving''}
% \vspace{-15pt}
\begin{abstract}

In this paper, we focus on the problem of  high-speed multi-robot navigation in cluttered spaces with tight spatial constraints. A key technical challenge that the literature today faces is to balance safety with agility, often trading one for the other (slowing down near a bottleneck to avoid a collision versus failing to swerve in time at high speeds) especially in car-like dynamic systems. The approach presented in this paper aims to solve this problem and belongs to the family of formation control methods where Ackermann-steering agents coordinate with each other when navigating tight spaces. Formation control techniques today, however, rely on rigid, pre-defined shapes and adaptations, initial conditions, environmental conditions, and vehicular dynamics, that are typically impractical to maintain at high speeds. For \textit{e.g.} agents going through a small gap can ONLY first form a square that eventually transforms into a line using analytical equations, and then back into a square. Our approach, on the other hand, draws inspiration from a smoothly flowing droplet of water that adapts its shape continuously and dynamically through whatever space its flowing without slowing down. We propose a hierarchical algorithm that, at the high level, trains a reinforcement learning policy to produce a fast, dynamic ``blob-like'' elliptical patch that can morph its shape in real-time and can squeeze, stretch, and orient itself fluidly around obstacles and through tight spaces. At the lower level, the agents or robots are optimized via distributed model predictive control (DMPC) to stay inside the patch (which acts as a soft forcing function for the agents) and not collide with each other. We evaluate our approach, \model, using simulated F1TENTH racing cars and show that \model surpasses the baselines and achieves a completion rate of 89\% at $7.73$ m/s across $N{=}1$ to $N{=}6$ robots in 14 different scenarios averaged over 20 trials each, while all baselines either fail to complete the track or succeed at extremely low speeds ($0.07$--$1.79$ m/s, ~40\% completion rate). Supplementary code, videos, and additional results can be found at \href{https://autonavbot.github.io/Fastfunnels/}{https://autonavbot.github.io/Fastfunnels/.}

\end{abstract}

\section{Introduction}  \label{sec:Intro}

Agile (defined as high-speed mobility in narrow spaces) multi-robot autonomous navigation in tight, cluttered environments has become essential for applications such as search and rescue, warehouse logistics, and exploration in dense environments\cite{alonso2017shape,zhou2018agile}. The fundamental hardness of this problem is the conflict between continuously making progress towards the goal and satisfying dynamic environmental constraints at high speeds.  To solve this problem, there has been significant interest in developing decentralized solutions that typically frame multi-agent navigation as a joint trajectory optimization problem~\cite{chandra2026multirobotnavigationsocialminigames}. However, solving joint optimization problems in real time is very computationally expensive~\cite{lidard2024blendingdatadrivenpriorsdynamic, hu2024playsfirstoptimizingorder, chandra2025deadlock}
% ; the high-dimensional joint state space, combined with non-convex safety constraints in cluttered environments, creates an optimization landscape dominated by local minima. Consequently, traditional solvers struggle to have a consistent balance of solutions that satisfy both safety margins and high-speed dynamic constraints
% Decentralized methods tackle the challenge by assigning independent goals to each agent~\cite{inbook}, \cite{ames2019controlbarrierfunctionstheory}. 
Literature in multi-robot coordination suggests that part of the computational complexity can be mitigated by leveraging \textit{adaptive formation control} to improve spatial coordination~\cite{10933518, park2023formation, quan2023formation, deng2025coordinatedmultirobotnavigationformation}. By operating with a shared spatial consensus, a team of robots can collectively adapt to environmental constraints as a unified entity, preventing deadlocks, bottlenecks, and team fragmentation (e.g., stragglers) that otherwise occur when individual agents greedily optimize their own localized trajectories. 

% To establish the necessary coordination and bypass computational bottlenecks, formation control emerges as a mathematically viable strategy~\cite{deng2025subteamingadaptiveformationcontrol}. 
% It solves the intractable optimization problem through spatial abstraction and dimensionality reduction. By treating $N$ robots as one single shape, you drastically reduce the number of variables the computer has to calculate, which makes the math fast enough to run in real-time (allowing for high speeds). 

However, current formation control methods~\cite{10933518, park2023formation, quan2023formation} rely on rigid formations or ad hoc/rule-based protocols for coordination, which is overly conservative and unsafe in high-speed multi-agent navigation especially in narrow spaces. For example, if the passage geometry is narrower than the formation's fixed shape, boundary violations can occur and the team loses its safety. To mitigate this, rule-based methods and discrete shape libraries enforce a narrower, predefined shape (\textit{e.g.}, switching from a wedge shape to a line shape) before entering a tight corridor. However, predefined shapes are overly conservative, as they often force the team to drastically decelerate or halt entirely before the bottleneck, destroying the fluid momentum required for high-speed mobility. 
% Our proposed approach belongs to the category of formation-based control for multi-robot navigation.
% Thus, traditional methods force a binary compromise: operate safely but overly conservatively, or operate quickly but unsafely. \rc{double check - i think you explained the overly conservative limitation twice?}

\begin{figure*}[t]
  \centering
  \includegraphics[width=\textwidth]{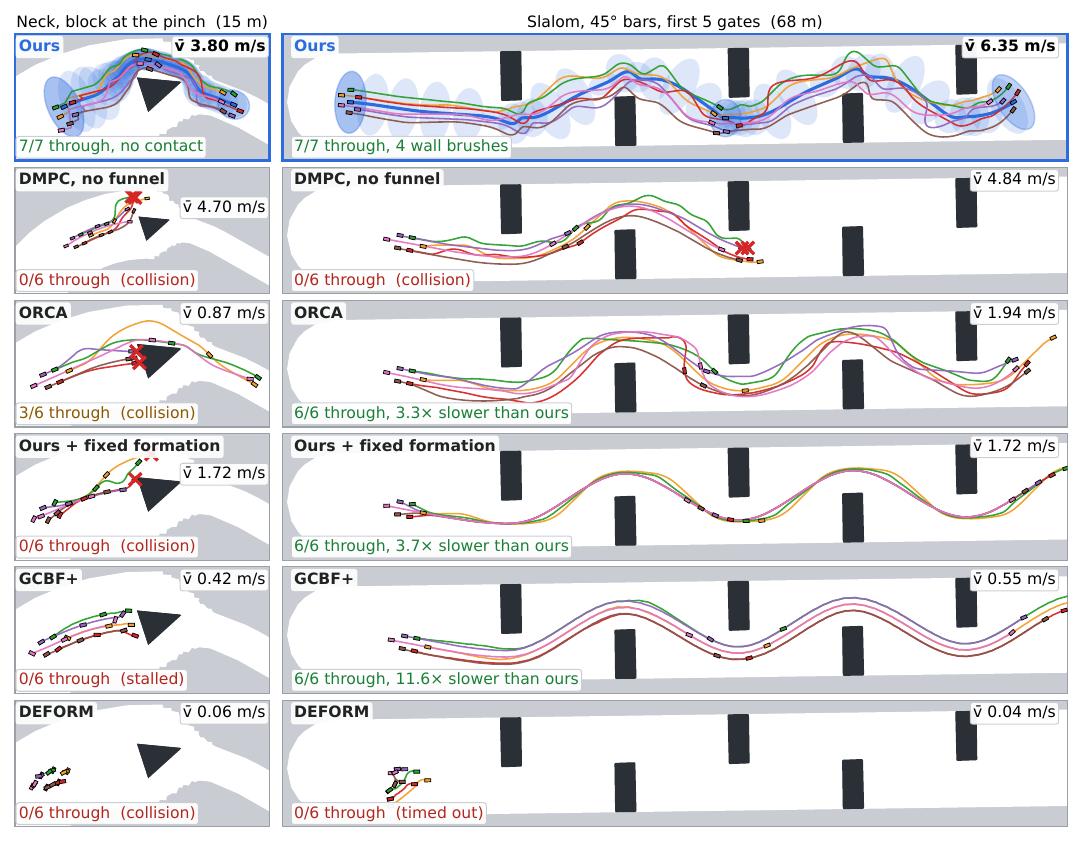}
  \vspace{-10pt}
  \caption{Results for a N = 6 shown for 2 (out of 14) benchmark scenes. Each scene is averaged over 20 trials. Remaining scenarios and results can be found in the Appendix. The \textcolor{ff}{\textbf{elliptical funnel patch}} dynamically adapts its shape and orientation at every control step acting as a soft forcing function to guide the agents fluidly through bottlenecks and narrow gaps at high speed, much like how water flows around rocks and bends. We have tested with $N=6$ robots but \model can scale up to 12 robots as well (beyond $N=12$ not tested).}
  % Ours show 7 agents, as $N$ does not include the patch car(leader) and its always $1+N$ agents in the scene}
  \label{fig:qualitative}
  \vspace{-15pt}
\end{figure*}

\noindent\textbf{Novelty:} This work extends the current state-of-the-art in formation control~\cite{10933518}: going \textit{beyond pre-defined} formations for unicycle dynamics and implementing real-time, \textit{continuously adapting}, high-speed formations for Ackermann steering agents. Our approach also differs in the way the multi-robot system moves through narrow spaces--unlike the rigid motion of prior methods~\cite{10933518, park2023formation,quan2023formation}, our system mimics the smoothness and agility of water droplets as it flows around bends and corners. The core technical insight of our underlying algorithm is to separate multi-agent interaction, safety, and fluid agility---because integrating all three together in a multi-robot system is extremely challenging and is the primary reason for the rigidity and conservativeness of prior methods. By delegating fluid agility to a high-level object (a virtual ``blob-like'' deformable elliptical funnel patch), the multi-robot system only needs to satisfy basic safety and containment constraints---something which is far easier.

\noindent\textbf{Main Contributions:} This paper proposes a novel hierarchical algorithm that enables high-speed adaptive formation control for F1TENTH vehicles with real car dynamics with Ackermann steering. A reinforcement learning leader policy guides a team of follower distributed MPC agents. Formally we contribute:
\begin{enumerate}
    \item A reinforcement learning policy that trains a ``blob-like'' deformable elliptical patch that can morph its shape in real-time and can squeeze, stretch, and orient itself fluidly (like a water droplet) around obstacles and through tight spaces. (Section~\ref{sec:leadermdp}).
    % \vspace{-15pt}
    \item A real-time distributed model predictive control (DMPC) optimization solver for each ackermann-steering agents such that they stay inside the patch and not collide with each other (Section~\ref{sec:Followerphase}). The patch serves as a spatial guide (or a soft forcing function) for the agents, instead of requiring the agents to assume a specific shape as is done in typical formation control methods, which is what allows the system (patch + agents) to continuously change its shape at high speeds. 
    % {\color{red}{\item (\textbf{Agent policy}) A robust containment policy trained against the dynamic patch policy, enabling agents to safely maintain their position inside the continuously morphing patch under non-linear, single-track kinematic constraints (Section~\ref{sec:followermdp}).}}
    % \vspace{-5pt}
    \item To test the limits of \model, we chose the state of the art F1TENTH racing simulator that has extremely fast racing car agents (up to 20m/s). We compared with the state of the art formation control baseline (DEFORM~\cite{10933518}), velocity obstacles-based collision avoidance~\cite{inbook}, multi-agent graph-based method with control barrier functions~\cite{berducci2023learningadaptivesafetymultiagent}, and a multi-agent reinforcement learning baseline (Section~\ref{sec:experiments}).
    % \vspace{-15pt}
    \item Real world experiments are conducted on 1/7th scaled Traxxas UDR vehicles which use Ackermann steering (Figure~\ref{fig:cover}). They are deployed with lidar sensing on a cluttered narrow corridor with static obstacles along the way. 
\end{enumerate}

\noindent \textbf{Main Result.} Our main result is that \model completes all 20 trials of 14 scenarios at an average of 40\% higher completion rate while being 7.69$\times$ faster than the fastest baseline. Also, even when \model slows down around the bottlenecks, it is \textit{still} 1.2$\times$ faster than DMPC, 2.3$\times$ faster than ORCA and 6.5$\times$ than GCBF+.

\section{Related Work}
We focus on various approaches to multi-robot navigation and divide our discussion along three dimensions: $(i)$ methods that focus on the problem of agile navigation in tight, narrow, and cluttered spaces, $(ii)$ methods that attack this problem using formation control in particular, and $(iii)$ methods that use advanced coordination strategies (such as game theory, graph search, and reinforcement learning). We keep our discussion brief particularly on the last dimension, focusing more on the role that each algorithmic subclass plays in the problem studied in this paper. For a more comprehensive survey of multi-robot navigation methods in cluttered environments, we refer the reader to the review conducted by Chandra et al.~\cite{chandra2026multirobotnavigationsocialminigames}.

% Our work relates to recent advances in multi-robot navigation, coordinated navigation, game-theoretic
% motion planning, and multi-agent path finding.

\subsection{Multi-Robot Navigation in Narrow and Tight Spaces}

The reason why many multi-robot navigation systems slow down in narrow spaces is that they enter a deadlock in trying to preserve safety~\cite{grover2020does, impc, grover2016before}. One family of methods~\cite{wang2017safety, impc, zhou2017fast} aim to perturb the system of agents if they start to slow down or enter a deadlock using the right hand rule (agents to the right receive higher priority). These strategies are effective in structured settings, but the imposed pre-determined ordering limits their generalizability and robustness and many cannot generalize to more than $3$ agents. Furthermore, these methods resolve deadlocks after the fact, which is already too late. Another line of research \textit{prevents} deadlock by identifying and mitigate potential deadlocks, even before they happen, such as in~\cite{chandra2025deadlock, lee2025merry, csenbacslar2024dream}.

A third family of approaches uses multi-agent pathfinding to resolve conflicts at narrow gaps~\cite{dergachev2021distributed, chandra2023socialmapf}. While MAPF in general produces discrete paths, extensions to continuous time and kinodynamic motion improve realism for physical robots, but they also increase the dimensionality and complexity of the search space~\cite{andreychuk2022ccbs,honig2018trajectory}. A fourth family of approaches uses multi-agent reinforcement learning (MARL). Multi-Agent Deep Q-Networks (MADQN)~\cite{orr2023multi} extend Deep Q-Networks to multi-agent settings. Value-based methods suffer from scalability issues due to joint action space. On the other hand, policy-based methods like MADDPG~\cite{lowe2017multi} enables each agent to learns its own deterministic or stochastic policy. A fundamental challenge in MARL is to achieve Nash equilibrium (NE) solutions where no agent can unilaterally improve their performance by deviating from their policy. Finally, dynamic game-theoretic solvers compute optimal solutions for multi-agent interactions in narrow and tight spaces~\cite{wang2020game}, but are too slow, computationally intractable for many scenarios, and require knowledge of the objective functions of other agents~\cite{laine2023computationoflocal, chandra2025mairl}. As noted by Zhou et al.~\cite{impc,zhou2017fast}, developing provably safe, deadlock-free, and fast algorithms to this problem is still an open problem. While many papers discussed above have made progress to some degree on the first and second axes (safe and deadlock-free), our work pushes the envelope towards the second and third axes (deadlock-free and fast).

\subsection{Multi-Robot Navigation using Formation-Based Control}

Multi-robot navigation has historically relied on foundational formation control strategies such as leader-follower, virtual structure, potential-field, and behavior-based approaches~\cite{oh2015survey, chen2009leader, duran2010adaptive, sabattini2011arbitrarily, xue2010formation, wang2020shape, morgan2016swarm, alonsomora2016distributed}. These methods established core ideas for coordinated motion, but they are often sensitive to obstacle density and to the geometric constraints imposed by narrow passages or cluttered environments. 

More recent work has extended formation control to support adapting and reconfiguring the formations in more challenging settings \cite{park2023formation,quan2023formation, 10933518}. However, these methods can still struggle to preserve a user-specified formation while adapting to tight environments, especially when the configuration must be reshaped on the fly. The DEFORM algorithm~\cite{10933518} dynamically adjusts a multi-robot system's shape to navigate unknown, obstacle-dense environments safely. It operates by using the clearance width to select a particular shape from a predefined shape library. The current available shapes include square, line, and wedge. It takes them several seconds to shift between shapes and can operate under holonomic conditions at a maximum speed of 0.2 m/s. 

% \begin{figure}[t]
%   \centering
%   \includegraphics[width=1.0\columnwidth]{images/fig_adapt_column.pdf}
%   \caption{Continuous adaptation (\model) vs. template switching (DEFORM). (\textit{top}) DEFORM~\cite{10933518} takes 25 template switches and 172 seconds to pass the track in single file. \textit{(middle, bottom)} \model passes the same track and another track in 1.5 seconds at 10m/s by widening the funnel to $1.9$ meters between the walls and closing it to $1.0$ meters at the block.}
%   % Bottom: a block, then a block in a narrowing bend the funnel shrinks in both axes and the funnel car eases from 9.9 to 6.9\,m/s. Tile labels: time since scene entry, and $b$ (m) for ours.}
%   \label{fig:adaptation}
%   \vspace{-15pt}
% \end{figure}

Park and Lee~\cite{park2023formation}  propose a particle swarm optimization-based formation controller to conduct formation reconfiguration. The algorithm combines a receding-horizon predictive controller with the particle swarm optimizer. At each time step, the global optimizer explores candidate control inputs across the state space to bypass local minima, while the MPC framework enforces kinematic constraints and guarantees real-time, collision-free trajectory execution. However, this controller utilizes a reduced search space, which leads to the absence of feasible solutions especially in cluttered narrow spaces (as evidenced by this work's results which have only been demonstrated in open spaces with no obstacles). Quan et al.~\cite{quan2023formation} presents an optimization-based, distributed trajectory planning framework designed for autonomous multi-UAV swarms navigating cluttered environments in formation. The framework relies on rigid target positions or fixed displacement vectors relative to a formation center. In obstacle-dense environments, rigid shapes present critical challenges.

DEFORM~\cite{10933518}, Park and Lee's approach~\cite{park2023formation}, and Quan's method~\cite{quan2023formation} are current adaptive formation approaches (with DEFORM being the state of the art). \model is a leap forwards from these methods by enabling continuous and dynamic formation adaption at high speeds on real car-like dynamics with Ackermann steering, \textit{without needing a library of pre-defined shapes}.

\section{Preliminaries}
% \vspace{2pt}
% \rc{check this place in the latex file for flow structure}

\begin{comment}
\section{Offline Reinforcement Learning for High-speed Multi-Robot Navigation}

<Intro text> as it appears already.

\subsection{Problem Formulation and Assumptions}
- An MDP is generally defined as ...
- typical objective in MDP is max_\theta J(.)
- our problem is formulated as heirarchical...Leader.. follower - solved sequentially. 
- briefly define what the leader and followers are. For the followers, mention if they are point-like holonomic agents with no shape and no dynamics or are they car-like objects with real-world dynamics.
- explain the scenario more explicitly. Mention narrow F1/10th track and the nature of the obstacles.
- our overall objective is max_\theta L(.) = J1 + J2, where J1 and J2 are explained in next couple of subsections.
- Assumptions (center line? any assumptions on dynamics? any assumptions on knowledge of other agents? communication? centralized vs decentralized?)

\subsection{Leader MDP}
- describe the tuple (state, obs, actions, etc.)
- describe the patch
- state the objective J1

\subsection{Follower MDP}
- what information is passed from leader to follower? are the multiple N followers decentralized?
- describe the tuple (state, obs, actions, etc.)
- state the objective J2
\end{comment}

We consider the problem of high-speed multi-robot navigation through a highly constrained and cluttered environment.
% We propose a leader-follower paradigm in which a dynamic, safety-guaranteed region bounds the workspace of a secondary learning fleet.

\subsection{Problem Formulation}

We cast high-speed multi-robot navigation as a {hierarchical} problem. A deformable elliptical patch object acts as the leader and is trained via reinforcement learning (\ref{sec:leadermdp}). $N$ robots, each of which is an independent follower, solve a distributed model predictive control (DMPC) problem in each control cycle using the leader as a soft forcing function to constrain the robots inside the elliptical patch (\ref{sec:Followerphase}). The leader trains a policy $\pi_\theta$ that outputs the actions for the patch. Once trained offline, the frozen policy ($\pi_{\theta^*}$) is used within the constraints of DMPC formulation used by the follower robots. 

At each control time step, the patch policy acts first via $\pi_{\theta^*}$, with the followers then solving the MPC problem and applying their first inputs before the simulator advances. Each follower robot estimates the patch's future shape, velocity and position using $\pi_{\theta^*}$, as well as shares its own planned trajectory with other robots. Each robot employs a geometrical half-plane constraint to avoid a congestion or a traffic jam and always helps them find a slot within the patch.

\subsection{Vehicle dynamics}
\label{sec:dynamics}
The leader patch as well as the follower robots are modeled using the single-track dynamic model \cite{Althoff2017CommonRoad} of the F1TENTH simulator~\cite{okelly2020f1tenth}, which includes tire slip angles, lateral tire forces, and yaw inertia rather than assuming instantaneous kinematic response. The state and control are defined as: 
\begin{equation}
\label{eqn:state}
 \xi_t^{\!\top}:=\begin{bmatrix}x_t & y_t &  \delta_t & v_t & \psi_t & \dot\psi_t & \beta_t\end{bmatrix}
\end{equation}
\vspace{1pt}
\begin{equation}
 u_t:=\begin{bmatrix}\alpha_t\\ \dot\delta_t\end{bmatrix}
=\begin{bmatrix}\text{longitudinal acceleration}\\ \text{front steering rate}\end{bmatrix}
\end{equation}
where $x_t$, $y_t$ are the global spatial positions of the center of gravity, $\delta_t$ is the front steering angle, $v_t$ is speed, $\psi_t$ is the yaw angle, $\dot\psi_t$ is the yaw rate, and $\beta_t$ is the slip angle. The vehicle dynamics are represented in state space as: (additional details on vehicle dynamics are in the Appendix). 
% \rc{can you add back the details on $f()$, $A$ and $B$}
% \begin{equation}
% \label{eq:stmodel}
% \dot{ \xi_t}= f( \xi_t, u_t)= A( \xi)+ B\, u
% \end{equation}
\begin{equation}
\label{eq:stmodel}
\dot{ \xi_t}= f( \xi_t, u_t)
\end{equation}
where $\xi_t$ is the current system state and $u_t$ the vehicle input at the current system state. All agents share the dynamics of Eq.~\ref{eq:stmodel} with input $(\alpha_t, \dot\delta_t)$.
% \begin{equation}
% \label{eq:stmodel}
%  A( \xi)=\begin{bmatrix}
% v\cos(\psi+\beta)\\ v\sin(\psi+\beta)\\ 0\\ 0\\ \dot\psi\\ \ddot\psi( \xi)
% \end{bmatrix},
% \end{equation}
% \begin{equation}
%  B^{\!\top}=\begin{bmatrix}0&0&0&1&0&0&0\\ 0&0&1&0&0&0&0\end{bmatrix}
% % \end{bmatrix}
% \end{equation}
% Refer 
% where $\ddot\psi( x)$ \& $\dot\beta( x)$ are the lateral tyre-force terms of the single-track model, linear in $\delta$, $\beta$, \& $\dot\psi/v$ with cornering stiffnesses $C_{Sf},C_{Sr}$, friction $\mu$, and the load transfer $g,l_r\mp\alpha h$. \rc{be careful when making edits, especially when removing stuff. I dont see these tire forces appear anywhere in the above equations. If you're removing stuff but there's a reference here in the text, make sure to say that it's given in the appendix.}

% \rc{where are J patch and J agent appearing in later sections? You need to connect back to these two terms.}
% \vspace{-5pt}
\subsection{Assumptions}
\label{subsec: assumptions}
\emph{(i) Map:} the track occupancy grid and centerline are known at initialization only and there are no dynamic obstacles beyond the
robots themselves. \emph{(ii) Homogeneous dynamics:} all $N$ robots share the same single-track dynamics.
\emph{(iii) Spawn condition:} robots spawn inside the patch. The patch is trained to admit at least $N$ non-overlapping footprints. 
\emph{(iv) Observability and execution:} \label{assumption:four} each robot observes its own ego state, the current patch parameters $(x_t,y_t,a_t,b_t)$, the patch velocity (${v_t}$), and and the planned trajectories of the other agents. Execution is decentralized and each robot runs its own controller. \textit{(v) Safety:} Dynamics and actuator bounds are hard constraints, whereas containment is penalized and separation admits penalized slack. These relaxations permit recovery from containment or separation violations but do not guarantee collision avoidance, forward invariance, or feasibility of the remaining hard constraints.

\section{\model: A Hierarchical Approach}

In this section, we present the formulation of the hierarchical algorithm consisting of a Leader--a reinforcement learning-based blob-like elliptical patch--that continuously morphs and adapts to the environment and multiple Followers that consist of robots that solve a distributed MPC optimization.

\subsection{Leader (Patch): learned offline}
\label{sec:leadermdp}
The leader is modeled as a partially observable Markov decision process (POMDP). In general, a POMDP is a tuple, $\mathcal{M} = \{\mathcal{S}, \mathcal{A}, \mathcal{P}, \mathcal{Z}, r, \gamma\}$, 
where $s_t\in\mathcal{S}$ is the state, $u_t\in\mathcal{A}$ is the action, $\mathcal{P}(s'_{t+1}\mid s_t,u_t)$ the transition kernel, $\mathcal{Z}: \mathcal{S} \to \mathcal{O}$ is the observation function, $r_t(s,u)$ the reward, and $\gamma \in (0,1)$ the discount factor. At time $t$ the agent receives $o_t = \mathcal{Z}(s_t)$, draws ${u}_t \sim \pi_\theta(\cdot \mid o_t)$, collects
$r_t(s_t, {u}_t)$, and the state evolves as $s_{t+1} \sim \mathcal{P}(\cdot \mid s_t, {u}_t)$. The reinforcement learning (RL) objective is to learn a policy $\pi_\theta$, using its trainable parameters $\theta$ that maximizes the expected discounted return $J(\theta)$. The leader MDP is defined as the tuple,

% \vspace{-10pt}
{\small\begin{equation}
  \mathcal{M}^{\mathrm{patch}} :=
  \bigl\{\mathcal{S}^{\mathrm{patch}},\;
          \mathcal{O}^{\mathrm{patch}},\;
          \mathcal{A}^{\mathrm{patch}},\;
          \mathcal{P}^{\mathrm{patch}},\;
          r^{\mathrm{patch}},\;
          \gamma\bigr\},
    \label{eq:leadermdp}
\end{equation}}

where the transition $\mathcal{P}^{\text{patch}}$ follows the single-track
model of Eq.~(3) as implemented in the F1TENTH gym (Appendix). The leader
state $s_t = (\xi^\text{patch}_t, a_t, b_t) \in \mathcal{S}^{\text{patch}}$ comprises
the single-track state $\xi^\text{patch}_t$ of the patch car (Eq.~(1)) and the
semi-axes $(a_t, b_t)$ of the elliptical patch. The policy does not observe
$s_t$ directly; it receives
$o_t = \mathcal{Z}^{\text{patch}}(s_t) \in \mathcal{O}^{\text{patch}} \subseteq \mathbb{R}^{115}$,
formed by concatenating seven Frenet-frame scalars with a sub-sampled
LiDAR scan of 108 beams. At each timestep the policy samples an action
tuple.
% where $\mathcal{S}^{\mathrm{patch}}$ is the state space,
% $\mathcal{O}^{\mathrm{patch}}$ is the observation space,
% $\mathcal{A}^{\mathrm{patch}}$ is the allowed action space, $\mathcal{P}^{\mathrm{patch}}(s'_t\mid s_t,u_t)$ is the transition which follows the dynamic model implemented in the
% F1TENTH gym (more details in the Appendix), $r^{\mathrm{patch}}_t:\mathcal{S}^{\mathrm{patch}}\!\times\!\mathcal{A}^{\mathrm{patch}}\!\to\!\mathbb{R}$
% the reward, and $\gamma\in(0,1)$ the discount factor.
% %% rc1: observation space now appears in the tuple
% The leader state $s_t=( \xi^{\mathcal P}_t,\,a_t,\,b_t)\in\mathcal S^{\mathrm{patch}}$
% represents the single-track state of the patch using Eq.~\eqref{eqn:state}, where $(a_t, b_t)$ denote the elliptical patch's semi-axes' lengths. The policy does not observe $s_t$ directly; instead, it receives an observation $o^{\mathrm{patch}}_t \in \mathcal{O}^{\mathrm{patch}} \in \mathbb{R}^{115}$
% formed by concatenating seven Frenet-frame scalars with a sub-sampled LiDAR scan. At each timestep, the policy samples an action tuple, 

\begin{equation}
\label{eq:actiontuple}
u^{\mathrm{patch}}_t = \bigl(\bar\delta_t,\;
        \bar v_t,\; a_t,\; b_t\bigr)
        \;\in\;\mathcal{A}^{\mathrm{patch}}
\end{equation}
% \begin{equation}
  % u^{\mathrm{patch}}_t = \bigl(\delta^{\mathrm{cmd}}_t,\;
  %       v^{\mathrm{cmd}}_t,\; a^{\mathrm{cmd}}_t,\; b^{\mathrm{cmd}}_t\bigr)
  %       \;\in\;\mathcal{A}^{\mathrm{patch}},
  % \label{eq:patch_action}
% % \end{equation}
% \rc{youre using u or a for patch action? also you said youre using the same dynamics for vehicle and patch, but vehicle action is different from patch actions. Equation 2 is different from equation 5}
% Speed and steering-angle setpoints are tracked by low-level controllers that produce the acceleration and steering-rate inputs of the model Sec. ~\ref{sec:dynamics}. 

The leader's policy emits speed ($\bar v_t$), steering ($\bar\delta_t$), setpoints and funnel axes $a_t, b_t$, and not $(\alpha_t, \dot\delta_t)$ directly because the controllers do different jobs, the patch policy additionally commands the funnel's shape, which is not a vehicle quantity. $\bar\delta_t$ is the steering-angle setpoint, $\bar v_t$ is the speed setpoint, and
$(a_t, b_t)$ are the commanded ellipse
semi-axes. For the patch dynamics, the driving commands $(\bar\delta_t, \bar v_t)$ are what produces the steering-rate and acceleration which are the actual inputs of the single-track model Eq .~\eqref{eq:stmodel}; the shape commands $(a_t, b_t)$ update the ellipse through the rate-limited integrator. Exact equations and limits are moved to Appendix. The per-step reward decomposes into five terms:
% \vspace{-3pt}

{\small\begin{equation}
  r^{\mathrm{patch}}_t \;=\;
       r^{\mathrm{prog}}_t
    \;-\; r^{\mathrm{track}}_t
    \;-\; r^{\mathrm{ctrl}}_t
    \;+\; r^{\mathrm{shape}}_t
    \;-\; r^{\mathrm{safe}}_t.
  \label{eq:patch_reward}
\end{equation}}

The progress term $r^{\mathrm{prog}}_t$ rewards forward motion along the centerline; the tracking penalty $r^{\mathrm{track}}_t$ discourages drift from centerline ($\mathcal{C}$); the control penalty $r^{\mathrm{ctrl}}_t$ penalizes aggressive steering and yaw rates at high speed; the shape reward $r^{\mathrm{shape}}_t$ encourages the ellipse to fill the local track width so followers have room to remain contained, and the safety penalty $r^{\mathrm{safe}}_t$ is a large negative signal triggered when the ellipse boundary intersects a wall or the car stalls. Exact functional forms and weights are deferred to Appendix. The leader's expected discounted return is then
% \vspace{-10pt}

\begin{equation}
  J^{\mathrm{patch}}(\theta)
    \;=\; \mathbb{E}_{\pi_\theta,\,\mathcal{P}^{\mathrm{patch}}}\!
          \left[\,\sum_{t=0}^{T} \gamma^{t}\,r^{\mathrm{patch}}_{t}
          \,\Bigm|\, s_0\right],
  \label{eq:J_patch}
\end{equation}

The leader then solves 
\begin{equation}
    \label{eq:phased_surrogates}
    \theta^* = \arg\max_\theta \mathcal{L}^{\mathrm{patch}}(\theta)
\end{equation}
where $\mathcal{L}^{\text{patch}}(\theta)$ is the PPO surrogate objective that runs gradient ascent, in place of $J^{\text{patch}}(\theta)$, as explained further in the discussion on our training routine in Section~\ref{sec:training_objective}.

% \rc{connect J patch here}
% \vspace{-15pt}
\subsection{Followers (robots): solved online using distributed MPC}
\label{sec:Followerphase}

Once the patch policy has been trained by maximizing the surrogate of
Eq.~\eqref{eq:ppo_loss_leader}, its weights $\theta^*$ are frozen. The
followers use the frozen policy within the MPC constraints. Each follower $i$ runs a nonlinear distributed MPC problem over $H$ steps each of length $\Delta\tau$ using the same MPC parameters. The team size can be chosen at deployment, without retraining the patch policy. Each follower at every control cycle minimizes a finite-horizon cost,
% \vspace{-15pt}

\begin{equation}
\min_{\xi^{i}_t,\;u^{i}_t}\sum_{k=0}^{H}\Big({J^{i,\mathrm{trk}}_t(k)}
+{J^{i,\mathrm{con}}_t(k)}
+{J^{i,\mathrm{int}}_t(k)}\Big)
\;+\;\sum_{k=0}^{H-1}{J^{i,\mathrm{sm}}_t(k)}
\label{eq:J_follower_new}
\end{equation}

subject to the vehicle dynamics of Eq.~\eqref{eq:stmodel} and the actuator limits. Additional detail on constraint are in the Appendix. The four cost terms are defined as follows: $J^{i,\mathrm{trk}}$ makes the agents to move at the predicted patch velocity and hold on its assigned slot position $q^{i}(k)$, $J^{i,\mathrm{con}}$ keeps it inside the predicted ellipse,  $J^{i,\mathrm{int}}$ keeps it clear of the other followers' planned trajectories, with the help of the shared position and velocity of each follower $j$, and
$J^{i,\mathrm{sm}}$ keeps the inputs small and smooth. The optimizer operates over a sequence of $H$ control steps, and only the first, $u^i_t(0)$, is applied to the vehicle with no low-level regulator in between, optimization is repeated at the next control tick. The follower's control law is therefore a function of what it knows at time step $t$,

\begin{equation}
   u^i_t(0)=\kappa\big(\mathcal I^i_t\big),
  \label{eq:kappa}
\end{equation}

where $\kappa$ returns the first input of the minimizer of
Eq.~\eqref{eq:J_follower_new} and $\mathcal I^i_{t}$ is the information set defined below. No follower parameter is learned--$\kappa$ is the same function for every follower and only its argument differs, which is why the number of followers $N$ is chosen at run time rather than at training time. Two ingredients make Eq.~\eqref{eq:J_follower_new} well posed inside a funnel that moves and deforms within a single horizon $H$, the follower must know where the funnel will be, and it must avoid the other follower without a central coordinator. We describe the information each follower holds and how it predicts the funnel (Sec.~\ref{sec:infoset}), then how the followers coordinate (Sec.~\ref{sec:coordination}).

% \vspace{-15pt}
\subsubsection{Information and funnel prediction $(\mathcal I^i_{t})$}
\label{sec:infoset}
Each follower $i$ uses its measured state, fixed slot index, the known centerline geometry, the current patch state and command, and the latest planned positions broadcast by other followers. It does not require the followers' measured states or future patch commands. Holding the current patch command constant over the horizon, it predicts the funnel pose, speed, and semi-axes using the vehicle dynamics and rate-limited shape update. This prediction lets followers anticipate funnel motion and deformation rather than track a frozen ellipse. Further details and mathematical descriptions are in the appendix.

% \indent{Control Constraints :} Two states are constrained with constant lower and upper bounds to ensure safe operation of the vehicle:
% \begin{equation}
%     \mathrm{v_{min}} \leq v \leq \mathrm{v_{max}}
% \end{equation}
% \begin{equation}
%     \mathrm{\delta_{{f}_{min}}} \leq \delta_f \leq \mathrm{\delta_{{f}_{max}}}
%     \qquad k=0,\dots,H
%     \vspace{-6pt}
% \end{equation}

% The control bounds are set to capture the actuator limits:
% \begin{equation}
%     \mathrm{-\alpha_{min}} \leq \alpha_k \leq \mathrm{\alpha_{max}}
% \end{equation}
% \begin{equation}
%     \mathrm{-\dot\delta_{{f}_{min}}} \leq \delta_f \leq \mathrm{\delta_{{f}_{max}}}
%     \qquad k=0,\dots,H -1
% \end{equation}
% $\delta_{i,k}$ is follower $i$'s steering angle at prediction step $k$ (a state, in radians), whereas $\dot\delta_{i,k}$ is its steering-rate input (in radians per second). The bounds $\delta_{\min},\delta_{\max}$ and $\dot\delta_{\min},\dot\delta_{\max}$ are the signed lower and upper limits on angle and rate, respectively.

\subsubsection{Coordination through trajectory exchange and separating half-spaces}
\label{sec:coordination}
Followers coordinate by exchanging plans rather than measured states. After each control step, follower $i$ broadcasts the position sequence ${\hat q^{i}_t(0:H)}$ that its minimizer visits, and every other follower plans against that sequence instead of against a static obstacle or a constant velocity extrapolation. The patch car enters as $j=0$ neighbor with the predicted funnel center of Sec.~\ref{sec:infoset} as its trajectory. For Every neighbor $j$ and every step $k$, a follower $i$ then constrains its planned position ${q^{i}_t(k)}$ to one side of a plane placed between ${\hat q^{\,i}_t(k)}$ and ${\hat q^{\,j}_t(k)}$:

\begin{equation}
{n^{ij}_t(k)}{}^{\!\top}\,{q^i_t}(k)\;\ge\;{d^{ij}_t(k)}-s^{ij}_t(k),
\qquad {s^{ij}_t(k)}\ge 0,
\label{eq:ocp_hp_new}
\end{equation}

where $n^{ij}_t(k)$ is the unit normal of the plane, $s^{ij}_t(k)$ is the slack variable on the separating plane, and ${d^{ij}_t(k)}$ is the clearance adjusted offset between followers the plane sits at half the required clearance from each, so each concedes the share it expects the other to concede, the patch car does not yield, so its plane is placed the full clearance from its path. Because the planes are coupled, each control tick runs two gauss seidel sweep: every follower in turn resolves  Eq.~(\ref{eq:J_follower_new}) against the others' latest plans and broadcasts its updated ${\hat q^{i}_t(0:H)}$. After the final sweep each follower applies ${u^{i}_t(0)}$ and replans at the next tick. Refer to the Appendix for more information the full information set, prediction notation, optimization constraints, separating-plane construction, and individual cost terms.
\section{Policy Training and Inference} 
\label{subsec:architectures}

% comment \rc{integrate the policy architecture subsections with the training pipeline subsection.}

Direct optimization of a joint policy that covers both patch scaling and multi-agent navigation presents severe non-stationarity challenges during training. To make optimization tractable, we exploit the natural asymmetric leader-follower structure of the formulation. Because the leader's state trajectory is structurally independent of the follower's positions, we decompose the problem hierarchically. The patch scaling and routing policy $\pi_{\theta}$ is trained first in isolation (leader MDP). Once converged, this policy is frozen, rendering the environmental variations completely stationary. The follower's online solver $\kappa$ is then deployed with the learned leader behavior.

\subsection{Training Objective}
\label{sec:training_objective}
Patch policy is optimized with Proximal Policy Optimization (PPO)~\cite{schulman2017proximal}, chosen for its stability under the
dense, mixed-scale reward terms of the leader's reward. The objective we ultimately want to
maximize is the leader return $J^{\mathrm{patch}}(\theta)$. However, Eq.~\eqref{eq:J_patch} admits  an expectation over trajectories whose distribution itself depends on $\theta$ and does not admits a closed-form gradient. Therefore, we use a surrogate objective which we optimize via Eq.~{\eqref{eq:phased_surrogates}. 
% We solve Eq.~\eqref{eq:phased_surrogates} as a proxy for the maximizer of $J^{\text{patch}}(\theta)$, with $\epsilon = 0.2$, $c_1 = 0.5$, $c_2 = 10^{-3}$, discount $\gamma = 0.99$ and GAE parameter $\lambda = 0.95$. 

% $J^{\mathrm{patch}}$ does not admits a closed-form gradient. It is an expectation over trajectories whose distribution itself depends on
% the policy parameters. PPO sidesteps this by replacing each $J$ \rc{what do you mean by each J?}with a
% \emph{surrogate loss} $\mathcal{L}$ whose gradient matches the true
% policy gradient $\nabla J$ at the behaviour policy and remains a
% valid ascent direction within a clipped trust region around it.
% %% added: explicit two-phase decomposition so reader knows there is
% %% NO single combined surrogate L = L^patch + L^agent
% The surrogate is optimized:

%  % \vspace{-3pt}
% {\small\begin{equation}
%   \theta^* \;\approx\; \arg\max_{\theta}\;\mathcal{L}^{\mathrm{patch}}(\theta)
%   % \qquad
%   % \phi^*   \;\approx\; \arg\max_{\phi}\;\mathcal{L}^{\mathrm{agent}}(\phi)
%   % \;\Big|_{\theta\,=\,\theta^*},
%   \label{eq:phased_surrogates}
%   % \vspace{-6pt}
% \end{equation}}

% \noindent which approximates the maximizer of
% $L(\theta)$\rc{fix. What is L($\theta$)}. 

\paragraph{Leader objective.}
\label{sec:lobj}
%% added opening tying L^patch to J^patch
% The leader is a single-agent problem and $\mathcal{L}^{\mathrm{patch}}(\theta)$
% is t
The PPO surrogate for the leader's true return $J^{\mathrm{patch}}(\theta)$ has the following standard composite form (where the expectations are taken over trajectories collected by the behavior policy $\pi_{\theta_{\mathrm{old}}}$).

{\begin{equation}
  \mathcal{L}^{\mathrm{patch}}(\theta)
  = \underbrace{\mathcal{L}^{\mathrm{CLIP}}(\theta)}_{\text{clipped surrogate}}
  \;-\; \underbrace{c_{1}\,\mathcal{L}^{\mathrm{VF}}(\theta)}_{\text{value loss}}
  \;+\; \underbrace{c_{2}\,\mathcal{H}[\pi_\theta]}_{\text{entropy bonus}},
  \label{eq:ppo_loss_leader}
\end{equation}}
% \vspace{-2pt}

\noindent where $c_1 = 0.5$, $c_2 = 10^{-3}$. The clipped surrogate $\mathcal{L}^{\mathrm{CLIP}}$ maximizes the expected
advantage while clipping the probability ratio
$\pi_\theta / \pi_{\theta_{\mathrm{old}}}$
to $[1{-}\varepsilon,\,1{+}\varepsilon]$, preventing any single update
from moving the policy too far from the behaviour that collected the
data. The value loss $\mathcal{L}^{\mathrm{VF}}$ minimises the
mean-squared error between the critic's predicted state value and the
empirical return, providing accurate advantage estimates via
Generalised Advantage Estimation (GAE). The entropy bonus
$\mathcal{H}[\pi_\theta]$ discourages premature
convergence to a deterministic policy by rewarding output diversity.
%% added closing: the operational link to J^patch
Stochastic gradient ascent on Eq.~\eqref{eq:ppo_loss_leader} yields
$\theta^* \approx \arg\max_\theta J^{\mathrm{patch}}(\theta)$. This
$\theta^*$ is then frozen and supplied to the follower phase.

Selecting $c_2$ proved non-trivial: with $c_2 = 10^{-2}$ the patch
policy collapsed into a high-entropy spinning mode with an unbounded
yaw rate and zero forward progress, a failure that required adding an
explicit yaw-rate penalty to the reward before entropy could be
reduced. The final value $c_2 = 10^{-3}$ provides a mild exploration
bonus without destabilizing cornering. Patch policy use a linearly
decayed learning rate ($3{\times}10^{-4} \to 1{\times}10^{-4}$).

\subsection{Inference}
At deployment, we use the mean of each Gaussian action head rather than sampling, eliminating the stochasticity used during training to explore. The policy $\pi_{\theta^*}$ is loaded with its frozen \texttt{VecNormalize} statistics from leader-phase training. At every control tick ($\Delta\tau = 0.01\,\mathrm{s}$) as in the F1TENTH simulator time step,  the leader first produces $u^{\mathrm{patch}}_t$ from $o^{\mathrm{patch}}_t$, the patch geometry is updated, the updated patch state is broadcasted.
\section{Experiments and Results}
\label{sec:experiments}
% \rc{check Table 2 in DEFORM paper. They measure the time taken to go from shape 1 to shape 2 and from shape 2 to shape 3. Can we show a contrast? show that in the time that DEFORM takes to change 3 fixed shapes, FF changes shapes continuously.}

Our experiments answer three key questions: (i) Does \model perform continuously make progress towards the goal through narrow spaces without compromising on speed? (ii) How significant is the impact of the patch in its role as a soft forcing function for the agents? (iii) How does \model compare to existing multi-robot navigation methods (both formation control-based as well as ones that are not) in terms of success, agility, and adaptability? To answer all these questions, we conduct experiments in simulation and on physical robots. First, we discuss the simulation setup, followed by the baseline and comparison methodologies for evaluation. We present and evaluate the findings to show how \model improves navigation safety and efficiency. We report metrics over 14 different scenarios averaged across 20 trials each, evaluation \model at key bottlenecks, narrow tunnels, and narrow gaps.
% isolating the behavior under spatial constraint, where coordination strategies diverge and all observed failures occur.

% recent cmt - \rc{nice! - can you darken the walls as well as the patches - they're too transparent and not easily visible}
%comment-  \rc{what are the questions you aim to answer? Refer to how we write these questions in Section 5 in https://arxiv.org/pdf/2603.21142, for example.}

% \subsection{Experimental Setup}
% \label{sec:evaluation}
% We evaluate \model on the F1TENTH Gym to perform the training and communication between the subagents and the patchcar agent. The simulator uses a single-track dynamic model ("st"), RK4 integrator, with a control timestep of $\Delta t = 0.01,\text{s}$. The Gym environment is used for its built-in support for Ackermann car-like dynamics and efficient modeling of 2D static environments.
 % \vspace{1pt}
% \begin{itemize}
%     \item \textit{Simulation Environment:} The F1Tenth Gym environment is used for its built-in support for Ackermann car-like dynamics and efficient modeling of 2D static environments.
%     \item \textit{Policy Training:} Stable Baselines3 PPO, VecNormalize, and SubprocVecEnv (parallel workers via SuperSuit for Phase A), and logged using WandB.
%     \item \textit{Sub-policy Training:} No concurrent communication between patch and agent; the patch policy is frozen and called as a Python callable inside each worker.
% \end{itemize}
% 

\subsection{Scenarios and Simulation Environment}
We evaluate on 14 different scenarios that are generated using 4 base maps: $(i)$ \textit{Cluttered environment:} Multiple obstacles are being placed at the neck of the  hallway entrance, such that the agents must squeeze past to avoid collision. $(ii)$ \textit{Multiple Narrow corridor:} A set of 3 to 4 long, winding corridor that gradually narrows from $\text{9.25 m}$ wide to $\text{3.3m}$, where the patch as to deform to traverse through, $(iii)$ \textit{High-Speed Corners:} A sequence of sharp, high-curvature corners designed to induce aggressive steering demands near the maximum $\delta_{\max}$, and $(iv)$ \textit{Slalom Environment:} A long bars protruding from the wall which are present relatively close to each other making the team perform a zigzag movement, this shows that ours is proactive as it speeds and slows down based on the difficulty of the scenarios. A full list and description of the $14$ scenarios is in the Appendix. To test the limits of \model, we chose the state of the art F1TENTH racing simulator~\cite{okelly2020f1tenth} that has extremely fast racing car agents (up to 20 m/s) with car-like dynamics using Ackermann steering.
% \rc{mention the evaluation zone - whole track vs part of the track and from whih point to which point are methods evalauted in?}

% \subsection{Robot Model and Dynamics}

% Each agent is modeled as a 3-DOF planar rigid body (Akinermann car model) where the control inputs ($u_i$) are throttle/acceleration and steering angle ($\delta$), constrained as follows:
% $$
% \dot{x} = v \cos(\theta)
% $$
% $$
% \dot{y} = v \sin(\theta)
% $$
% $$
% \dot{\theta} = \frac{v}{L} \tan(\delta)
% $$
% where $L$ is the wheelbase, $v \in [0, v_{\max}]$, and the steering angle $\delta \in [-\delta_{\max}, \delta_{\max}]$. This model ensures the feasibility of the hard $\mathbf{G}_{\text{Feasibility}}$ constraint within the SE-MPC formulation. 

% \subsection{Baselines and Ablations}

% Three specialized scenarios are designed in the F1Tenth environment to evaluate the system's ability to maintain high-speed "fluid flow" while satisfying the primary constraint (containment) and the safety constraint (no collision).

% \begin{enumerate}
\subsection{Baselines and Ablations}
\label{subsec:baselines}
% \rc{can you add your previous RL based version as a ``RL baseline'' ? reviewers might complain that we do not have a multi-agent RL baseline. So add that baseline, we know it doesnt work so thats good. Also add holonomic deform.}

We compared with the state of the art formation control baseline (DEFORM~\cite{10933518}), velocity obstacles-based collision avoidance (ORCA~\cite{inbook}), multi-agent graph-based method with control barrier functions (GCBF+~\cite{zhang2025gcbf+}), a multi-agent reinforcement learning baseline (IPPO with parameter sharing).
% \rc{explain how this leader follow is different from yours}.
% \vspace{-pt}
% \begin{figure}[t]
%       \centering  \includegraphics[width=0.9\columnwidth,]{images/fig_speed_profile_vs_N_paper.pdf}
%         \caption{Completion rate versus team size $N$ on the 14 benchmark scenarios. \model{} holds 88\,\%$\rightarrow$81\,\% from $N{=}3$ to $6$ while its funnel car runs at 7.7\,m/s for every $N$. The one baseline that is also fast, DMPC (4.7\,m/s), collapses from 43\,\% to 15\,\% as soon as more cars share the corridor; every other method stays below 60\,\% and crawls at $\leq 1.8$\,m/s (ORCA 1.1, Leader-Follower 1.8, GCBF+ 0.3, DEFORM 0.1). Marker area is the number of agents that completed and line-end labels give the mean speed $ V$.And Ours stays both fast and reliable as the team grows.\rc{seems to show the same information as table 1 - i think it might be redundant. I liked the figure where you showed that the slowing down we are still faster than baselines image. But maybe just use one of the three and improve the visuals a bit}}
% % The speed through the obstacle scenes vs. time, for N = 3 to 6 (mean over the 14 benchmark scenes; a team that collides or stalls counts as stopped from then on;
% % every method senses with onboard lidar + the corridor centerline only; ours = the funnel car's speed, which sets the team's pace).\rc{discuss this figure explicitly in the analysis} \rc{same as fig 4 - i think we can just show for N=6 to save space, move rest to appendix.}}
%       \label{fig:scalingN}
%       \vspace{-5pt}
% \end{figure}
To isolate the contribution of each component of \model, we
evaluate ablated variants that appear alongside the baselines in Table~\ref{tab:quantitative}. In the DMPC baseline, the patch is removed and agents operate in a fully decentralized sense as the same the initial assumption~\ref{assumption:four}. \model with a fixed patch evaluates the role of an adaptive patch and determines the extent to which an \emph{adaptive} ellipse deformation is necessary for containment. Finally, \model with fixed formation employs a fixed team geometry (\textit{e.g.} a wedge formation) with a learned leader policy that continuously reshapes and re-speeds based on the scene particularly focusing on the speed and adaptation.

\begin{table}[t]
\centering
\caption{Quantitative results across varying team sizes. \model completes all 20 trials of 14 scenarios at an average of 40\% higher completion rate while being 7.69$\times$ faster than the fastest baseline. (See Section~\ref{sec:metrics} for more details).}
\label{tab:quantitative}
\small
\setlength{\tabcolsep}{3pt}
\renewcommand{\arraystretch}{1.35}
\resizebox{\columnwidth}{!}{%
\begin{tabular}{rccccccc}
\toprule
\multirow{2}{*}{Method} & \multicolumn{4}{c}{\makecell{Completion rate (${C_r}$ \%)}}
  & ${V_{min}}$ &${\hat{V}}$ & $T_\text{clear}$\\
\cmidrule(lr){2-5}
 & $N{=}3$ & $N{=}4$ & $N{=}5$ & $N{=}6$ & (m/s) & (m/s) & (s)\\
\midrule
ORCA \cite{inbook}
  & 57 & 54 & 46 & 48 & 1.59& 1.13 & 28.80\\
\rowcolor{gray!20}
GCBF+~\cite{zhang2025gcbf+}
  & 18 & 19 & 17 & 18 & 0.70& 0.26 & 167.60 \\
\textsc{Multi-agent RL (IPPO)~\cite{schulman2017proximal}}
  & \multicolumn{7}{c}{unable to turn around corners} \\
\rowcolor{gray!20}DEFORM (F1TENTH) \cite{10933518}
  & 0 & 0 & 0 & 0 & 0.07& 0.00 & -- \\
  DEFORM (TB3) \cite{10933518}
  & 14 & 14 & 7 & 7 & 0.16 & 0.06  & 147.60 \\
\midrule
 DMPC
  & 43& 21 & 7 & 15 & 4.95& 4.68 & 3.80 \\
  
\rowcolor{gray!20}\textsc{\model{} + fixed patch}
  & 14  & 14  & 16  & 15  & 8.96 & 6.07  & 1.75 \\
\textsc{\model{} + fixed formation}
  & 17 & 14 & 13 & 12 & 1.88& 1.79 & 82.70 \\
\rowcolor{cyan!20}
{\textbf{\model}+\textbf{DMPC} \textbf{(Ours)}}
  & \textbf{90} & \textbf{89} & \textbf{86} & \textbf{81} & 5.07& \textbf{7.73} & 8.36 \\
\bottomrule 
\end{tabular}}
\end{table}

\subsection{Evaluation Metrics}
\label{sec:metrics}
% We evaluate the performance of our method with every other method on the same benchmark scenes quantitatively with the following metrics.
We use the following metrics: $(i)$ Completion Rate (${C_r}$) measures the number of agents that reach the goal averaged across all 14 benchmarks, 20 trials each. Collisions and deadlocks are counted as failures.
% the fraction of agents that cleared the scene, where an agent counts as successful only if it crosses the scene's finish line without any collision, i.e.\ it never comes within $0.15$~m of a wall or obstacle nor within $0.35$~m
% (centre-to-centre) of another vehicle while inside the scene. We report it as a percentage,
% summed over all scenes of a scenario, and a value of $100\%$ means that the whole team traversed the scene untouched.
% $(ii)$ Cohesiveness ($C_t$) is the percentage of the time for which each agent spent outside the elliptical funnel patch over 20 trials for all the 14 scenarios. This metric only applies for our method as there is boundary to stay in for the other methods. 
$(ii)$ (${V_{min}}$) is the mean lowest speed of a system during a particular bottleneck, $(iii)$ Average Velocity ($\hat{V}$) is the mean speed of each cars' speed throughout the scenario until the team crosses the finish, and $(iv)$ Time-to-Clear ($T_{\text{clear}}$), the total duration taken by the agents to navigate through the narrow corridor, measured from the first agent's entry to the last agent's exit (or total steps when operating in whole-track mode). 

\begin{figure}[t]
  \centering
  \includegraphics[width=1.0\columnwidth]{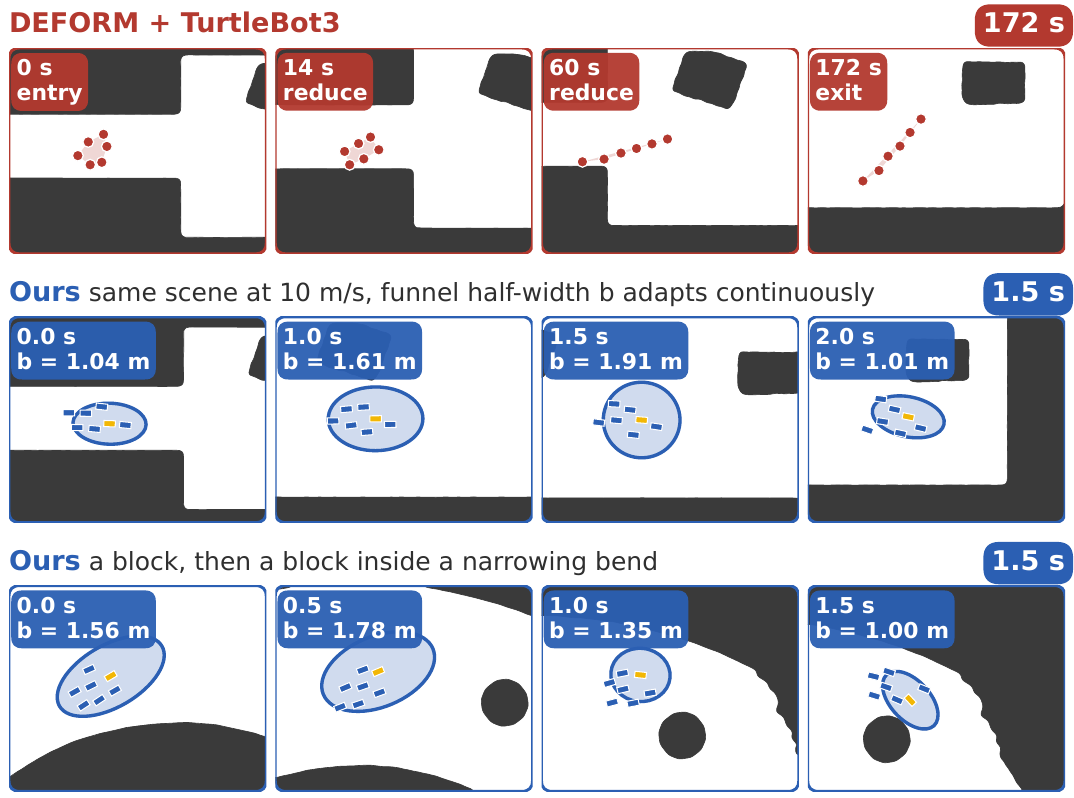}
  \caption{Continuous adaptation (\model) vs. template switching (DEFORM). (\textit{top}) DEFORM~\cite{10933518} takes 25 template switches and 172 seconds to pass the track in single file. \textit{(middle)} \model passes the same track and another track \textit{(bottom)} in 1.5 seconds at 10m/s by widening the funnel to $1.9$ meters between the walls and closing it to $1.0$ meters at the block.}
  % Bottom: a block, then a block in a narrowing bend the funnel shrinks in both axes and the funnel car eases from 9.9 to 6.9\,m/s. Tile labels: time since scene entry, and $b$ (m) for ours.}
  \label{fig:adaptation}
  \vspace{-10pt}
\end{figure}

% \subsection{Analysis and Discussion}
% \lipsum[3]
\subsection{Simulation Results and Analysis}
% {\rc D is not safety margin, this is not what D does}
The analysis below corresponds to Table~\ref{tab:quantitative} \& Figures~\ref{fig:qualitative} \& ~\ref{fig:speed}.
% \rc{re-check all values in this paragraph since you updated the table.} 
\paragraph{\model balances high speed with task completion.} 

\model achieves \textit{significantly higher} completion rates compared to the other baselines across different team sizes, while simultaneously maintaining a high average group speed of $7.73$ m/s, and this helps the team to clear the scene faster with an average time of $8.36$s across all the 14 scenarios which is is $3.34\times$ faster than ORCA ($28.8$s) and $19.5\times$ faster than GCBF+ agents ($167.6$s). \model also outperforms a state of the art adaptive formation based (Fig~\ref{fig:adaptation} method where it consumes $14$s to $60$s to switch between template/shapes as the planner as to compute a shape every time it something new or from a different angle, but ours does a continuous adaptation to the scene with a decision rate of $100$Hz (0.01 s) that is how quickly can the funnel reshape. Our main result is that \model completes all 20 trials of 14 scenarios at an average of 40\% higher completion rate while being 7.69$\times$ faster than the fastest baseline. Also, even when \model slows down around the bottlenecks, it is \textit{still} 1.2$\times$ faster than DMPC, 2.3$\times$ faster than ORCA and 6.5$\times$ than GCBF+ (Figure~\ref{fig:speed}).
% Compared to DMPC (no funnel), \model is $2.26\times$ slower because this average is calculated across all the successful completions and the DMPC completion rate is less so the avg time is also lesser this also reflects the benefit of the patch coordinating follower spacing rather than relying on reactive separation that increases the completion success. \rc{this feels like repetition - you are discussing dmpc in ablation}
% \begin{figure}[ht]
%   \centering
%   \includegraphics[width=1.0\columnwidth]{images/fig_speed_along_course_N6_column.pdf}
%   \caption{\rc{mention that \model is faster by $Z\times$ than the fastest baseline even when slowing down. Also, its hard to see the baseline curves because of all the red crosses clutter.}Speed drop across the scenario (N = 6): ours keeps 8 to 10 m/s through every block;
%   every baseline runs far slower and most collide at the first block they meet. \rc{discuss this figure explicitly in the analysis}. \rc{lets keep only one of the three, as we dont have space for 3 such plots}}
%   \label{fig:speeddrop}
%   % \vspace{-5pt}
% \end{figure}

\paragraph{Performance of baselines}
All the baselines including \model have the same initial observation of the scene, each car agent gets its own lidar scan and the raw centerline (local planning) of the scene. ORCA's poor performance is attributed to the fact that the velocity closest to its preferred velocity lies outside the velocity obstacle of neighbors and of nearby walls and obstacles, over a 2 second horizon, and points into dead block ahead, where the agents stall. GCBF+ suffers even lower completion rates as the control barrier function it uses forces the agents to act conservatively, which is why they move with an average speed of $0.26$ meters/second and decide to stop often rather than risk anything. DEFORM returns no control on any of the scenarios as can be observed in Figure~\ref{fig:qualitative}. This is because the approach is designed for built for differential drive Turtlebot3 dynamics. For fairness, we tested DEFORM with it's original settings in gazebo---while that improved the completion rate slightly, it still performed poorly. The multi-agent RL baseline does not scale as the shared policy PPO followers plateaued at $N=2$ robots. They failed to scale because each agent is rewarded only for keeping itself inside the funnel. Therefore, when several cars must share the same shrinking ellipse, the policy has no way to learn who yields and silently exits the patch in order to avoid inter collisions. The multi-agent reinforcement learning baseline only enables straight-line navigation and is unable to handle turning due to the above limitation.
% {\color{red}{ORCA succeeds at $N{=}2$ ($6.75$\,m/s, $aSR=eSR = 1.0$) but fails at $N{=}4$ ($aSR=0.0$): when four agents simultaneously enter the 3.3\,m corridor, the velocity-obstacle constraints become mutually infeasible, and the agents
% deadlock rather than negotiate passage. DEFORM stalls near zero speed ($0.01$\,m/s at $N{=}2$ with both success and safety at zero, indicating that its optimization cannot produce a feasible trajectory on the constrained geometry, the main reason for this is the kinematic model mismatch; the DEFORM controller is tuned for TurtleBot-like Twist behavior, but F1TENTH models Ackermann steering. Leader-Follower maintains high speed ($4.2$\,m/s at $N{=}2$) but achieves zero success across all $N$, with safety degrading from $0.50$ to $0.25$ as $N$ grows; the fixed-formation spacing cannot compress to fit multiple cars abreast through the corridor, so only the lead car clears while the rest stall or collide. ASRL reaches $aSR=0.80$ and 20\% success at $N{=}3$, but at only $0.74$\,m/s roughly $7.6\times$ slower than ours because its single ego-agent objective drives overly conservative behavior: the trained agent slows to avoid its fixed-policy neighbors rather than coordinating with them, preventing the full convoy from clearing
% together.}}
% \rc{you had made a column for "reasons why baseline fails" right? can you add a line each to each baseline summarizing what was the key reason for failure?}

\paragraph{Ablation study.}
When the learned patch is removed and \model is deployed with just the DMPC on each followers, the resulting baseline achieves higher lateral acceleration as shown in Table~\ref{tab:quantitative} but at the cost of safety (poor completion rates across all N), demonstrating that a learned funnel is necessary for reliable coordination. Also, replacing the adaptive patch with a fixed-shape ellipse (\textsc{\model + fixed patch}) reduces safety across all $N$: the rigid boundary intersects track walls and the static obstacles, forcing early termination. Together, these ablations confirm that adaptive deformation is necessary for the system's performance and success. Lastly, \model with fixed formation fails as expected because rigid formation behind the leader cannot adapt as needed, with the robots at the edge of the formation often clipping and running into the obstacles.

\begin{figure}[t]
      \centering  \includegraphics[width=1.0\columnwidth]{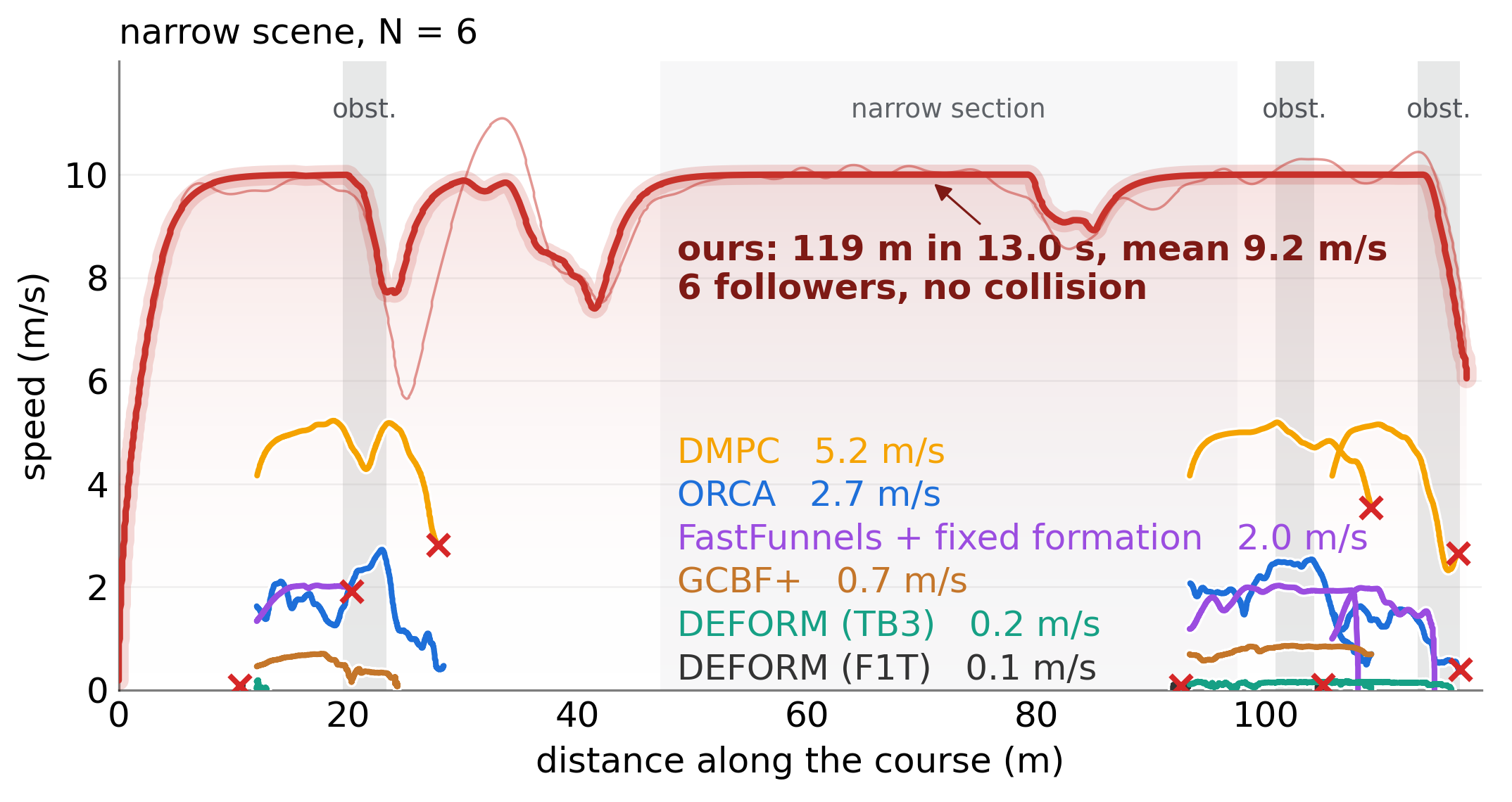}
      \caption{\model (red curve) at its lowest speed is still several orders of magnitude faster than the second-fastest approach.}
        % \caption{Speed along the narrow course (119 m, N = 6). Our funnel car (red; thin line: followers' mean) covers the course in 13.0 s at 9.2 m/s on average and slows only where the corridor demands it, at the obstacles (grey) yet even its slowest point, 7.4 m/s, is 42 \% faster than the fastest baseline ever gets (DMPC, 5.2 m/s peak) and 2.7× ORCA's peak. DEFORM(F1T) is inherently on the axis($0.07$-$0.13$ m/s and it barely moves and because of this it is hugging the wall and not very visible. The baselines, spawned 7 m before each obstacle, all stay below 5.3 m/s and every run ends in a collision (×); none reaches the narrow section.}

        % \rc{Figure 4 font needs to increase}}
% The speed through the obstacle scenes vs. time, for N = 3 to 6 (mean over the 14 benchmark scenes; a team that collides or stalls counts as stopped from then on;
% every method senses with onboard lidar + the corridor centerline only; ours = the funnel car's speed, which sets the team's pace).\rc{discuss this figure explicitly in the analysis} \rc{same as fig 4 - i think we can just show for N=6 to save space, move rest to appendix.}}
      \label{fig:speed}
      \vspace{-10pt}
\end{figure}

\subsection{Deployment using real robots}
We deployed \model on two 1/7-scale Traxxas UDR cars, with a 0.42m wheelbase, 6.1 kg, VESC drive, 2D lidar and a platform 1.3$\times$ larger and 1.6$\times$ heavier than a standard F1tenth simulator models (Figure~\ref{fig:cover}). 
The leader policy is retrained with the 1/7th Traxxas car plant and a 15m indoor corridor mapped with SLAM (1.6-3.1m wide, three cardboard obstacles), using a domain randomized pool of 350 corridor variants with lidar and pose noise. The trained policy runs in a single ROS2 node at 100Hz from the lidar scan and wheel odom, the odom is aligned to the track once at start. The cars traverse the corridor at up to 3.25m/s on average, with the funnel opening to 1.9m at straights and 1.1m at the obstacle gaps.The code and the checkpoints will be shared soon. 

\begin{figure}[t]
      \centering  \includegraphics[width=1.0\columnwidth,]{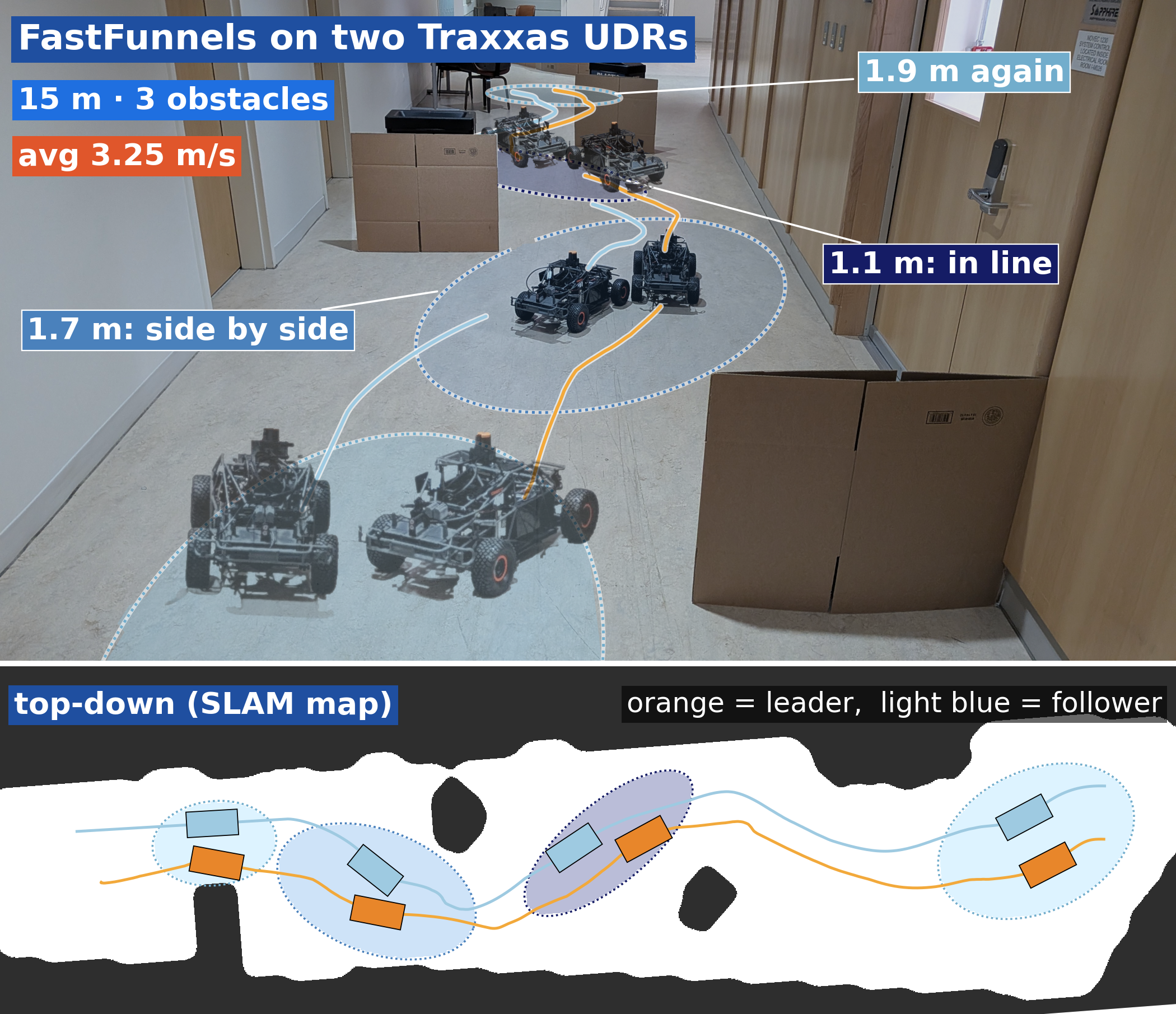}
      \caption {\model on two Traxxas UDRs (1/7th scale) in a $15$m narrow corridor with three obstacles. \textit{(top)} the funnel (\textcolor{ff}{\textbf{blue elliptical patch}}) is trained to morph (1.9 meters to 1.1 around the gap and then back to 1.9 meters in the open) its shape continuously and the followers are optimized to stay within the patch. 
      % Its width is decided by the leader from lidar alone: 1.9 m where the corridor is open, squeezed to 1.1 m at the box gap, re-opened after it without slowing down (avg 3.25 m/s).
      \textit{(bottom)}: top view.}
      % , the followers (light blue) re-form from side-by-side to in-line as the funnel narrows and back again, with no inter-agent negotiation or map, the funnel encodes the formation.}
      \label{fig:cover}
      \vspace{-10pt}
\end{figure}

\section{Conclusion, Limitations, and Future work}
% \lipsum[3]
In this work, we introduced a hierarchical framework for high-speed multi-robot navigation with car-like dynamics in cluttered environments with tight spatial constraints. Our main result is that \model completes all 20 trials of 14 scenarios at an average of 40\% higher completion rate while being 7.69$\times$ faster than the fastest baseline. Also, even when \model slows down around the bottlenecks, it is \textit{still} 1.2$\times$ faster than DMPC, 2.3$\times$ faster than ORCA and 6.5$\times$ than GCBF+ (Figure~\ref{fig:speed}). 

Currently, \model assumes that robots share information with each other and does not treat safety formally. These current limitations notwithstanding, more importantly, we provide several promising research directions and open problems as future work. For instance, exploring provable safety via control barrier functions, cutting off information sharing and making robots fully decentralized, extending to rough off-road terrains are some among several such directions for future work. 
{\footnotesize
\bibliographystyle{IEEEtran}
\bibliography{references, refs_smg, refs_mrn}
}% \bibliographystyle{corlabbrvnat}
\end{document}